\documentclass[conference]{IEEEtran}
\ifCLASSINFOpdf
\else
\fi
\AtBeginDocument{%
  }

\usepackage{microtype}
\usepackage{graphicx}
\usepackage{subfigure}
\usepackage{booktabs} % for professional tables

\usepackage{hyperref}

\usepackage{times}

\usepackage{fancyhdr}
\usepackage{color}
\usepackage{algorithm}
\usepackage{algorithmic}
\usepackage{multirow}
\usepackage{xspace}
\usepackage[commandnameprefix=ifneeded, final]{changes}
\usepackage{stfloats}

\usepackage{acronym}
\newacro{AI}{Artificial Intelligence}
\newacro{DAG}{directed acyclic graph}
\newacro{FL}{Federated Learning}
\newacro{ML}{Machine Learning}
\newacro{NN}{Neural Network}
\newacro{CNN}{Convolutional Neural Network}
\newacro{MPU}{memory protection unit}
\newacro{TMLaaS}{TinyML-as-a-Service}
\newacro{IoT}{Internet of Things}
\newacro{CBOR}{Concise Binary Object Representation}
\newacro{CoAP}{Constrained Application Protocol}
\newacro{IID} {independent and identically distributed} 
\newacro{MCU}{Microcontroller Unit}
\newacro{I$^2$C}{Inter-Integrated Circuit}
\newacro{IETF}{Internet Engineering Task Force}
\newacro{MLP}{Multi-Layer Perceptron} 
\newacro{ANN}{Artifical Neural Networks}
\newacro{TinyML}{Tiny Machine Learning}
\newacro{AIoT}{Artificial Intelligence of Things}
\newacro{ROC}{Receiver Operating Characteristic}
\newacro{AUC}{Area Under the Curve}
\newacro{WASN}{Wireless acoustic sensor network}
\newacro{ASN}{Acoustic sensor network}
\newacro{PTQ}{Post-Training Quantization}
\newacro{TPR}{True Positive Rate}
\newacro{FPR}{False Positive Rate}
\newacro{STFT}{Short-Time Fourier Transform}
\newacro{SNR}{Signal-to-noise ratio}
\newacro{DC}{Direct Current}
\newacro{OOM}{out-of-memory}
\newacro{NAS}{Neural Architecture Search}
\newacro{TVM}{Tensor Virtual Machine}
\newacro{DNN}{Deep Neural Network}
\newacro{MAC}{multiply–accumulate}
\newacro{RAM}{Random Access Memory}
\newacro{IR}{intermediate representation}
\newacro{SOTA}{state-of-the-art}
\newacro{DRAM}{Dynamic RAM}
\newacro{FPGA}{Field Programmable Gate Array}
\newacro{Conv1D}{1-D convolution}
\newacro{TCN}{Temporal Convolutional Network}
\newacro{RNN}{Recurrent Neural Network}
\newacro{SSM}{State-Space Model}
\newacro{GTA}{Global Temporal Aggregator}
\newacro{LSTM}{Long Short-Term Memory}
\newacro{GRU}{Gated Recurrent Unit}
\newacro{EEG}{electroencephalogram}
\usepackage{amsmath, amsfonts}
\usepackage[capitalise]{cleveref}

\newcommand{\methodName}{Tiny\emph{DéjàVu}\xspace}

\usepackage{balance}

\begin{document}

\title{
\footnotesize
\framebox[1.01\width]{\parbox{\dimexpr\linewidth-2\fboxsep-2\fboxrule}{If you cite this paper, please use the ISIoT 2026 reference: Z. Huang and E. Baccelli. \methodName: Smaller RAM and Faster Inference with Neural Networks on MCUs for Sensor Data Streams. in Proceedings of the 8th International Workshop on Intelligent Systems for the Internet of Things (ISIoT 2026).}}
 \ \\ \ \\ \ \\
\Huge\methodName: Smaller RAM and Faster Inference with Neural Networks on MCUs for Sensor Data Streams% with Always-On Microcontrollers
}

\author{
\iffalse
\IEEEauthorblockN{Zhaolan Huang}
\IEEEauthorblockA{Freie Universität Berlin, Germany \\
Email: zhaolan.huang@fu-berlin.de}
\and
\IEEEauthorblockN{Emmanuel Baccelli}
\IEEEauthorblockA{Inria Saclay, France}
\fi

\IEEEauthorblockN{Zhaolan Huang\IEEEauthorrefmark{1}, Emmanuel Baccelli\IEEEauthorrefmark{1}\IEEEauthorrefmark{2}}
\IEEEauthorblockA{\IEEEauthorrefmark{1} Freie Universit\"at Berlin, Germany \ \\
\IEEEauthorrefmark{2} Inria, France \ \\
Email: name.surname@fu-berlin.de
}
}


%Telephone: (800) 555--1212, Fax: (888) 555--1212}


\maketitle

\begin{abstract}
Examples of embedded intelligence include a wide variety of tiny neural networks used on-board wireless sensors and actuators, which are expected to continuously perform inference on time-series of the data they sense. In order to fit lifetime and energy consumption requirements when operating on battery, such hardware is exclusively based on microcontroller with as little memory as possible, e.g., 128 kB of RAM. In this context, optimizing data flows during inference across neural network layers becomes crucial. In this paper, we introduce a new framework, \methodName, and novel algorithms we designed to drastically reduce the RAM budget required by inference using various neural network models for sensor data time-series on typical microcontroller hardware. We publish the implementation of \methodName as open source, and we perform reproducible benchmarks on common microcontroller hardware (Arm Cortex-M). We show that \methodName can save up to 90\% of RAM usage with equal compute latency compared to prior work (StreamiNNC) on overlapping sliding window inputs.

\end{abstract}


\section{Introduction}

As \ac{AI} permeates all verticals, neural networks are deployed not only at the core of the network within large data centers, but also on much smaller end-user devices at the periphery of our distributed systems. The latter subdomain -- so-called \emph{edge \ac{AI}} -- includes use cases in which \emph{always-on} devices monitor the physical environment in their vicinity and perform inference on time series of the data they continuously sense, in real time. 

%embarking neural networks used to pre-process sensor data, and to take some decisions autonomously, such as sensed data classification, triggering network transmissions or actuators.

Popular examples include modern vocal interface use cases on wearable or portable devices with wake words (such as "\emph{Hey Google}" or equivalents). Precision agriculture or biodiversity monitoring use cases (such as TinyChirp~\cite{huang2024tinychirp}) incur continuous on-board classification on sensed data. Other use cases include compressed sensing and compressed communication over low-power networks, such as ~\cite{bernard2021compressed-sensing}.

When the power source of such devices is a battery, minimizing energy consumption and \ac{RAM} usage becomes crucial.
For this purpose, co-designing hardware and embedded software achieves great results.
In terms of hardware, minimizing energy consumption typically yields using microcontrollers as described in RFC7228~\cite{rfc7228}, coupled with specialized hardware acceleration, when available.

Embedded software running on such devices is programmed either \emph{bare-metal} at the register level, or coded against ultra-compact operating systems such as RIOT~\cite{baccelli2018riot}. The fields of \ac{TinyML} and edge \ac{AI} tackle challenges such as designing ultra-efficient neural network architectures and computation short-cuts to fit typically stringent resource constraints on microcontrollers. For instance: a total \ac{RAM} budget smaller than 50 kB, Flash memory smaller than 250 kB, and a single-core CPU running at 80 MHz, using various 32-bit architectures such as Arm Cortex-M, Espressif ESP32, or RISC-V 32-bit variants.

%For software, low-power \ac{TinyML}...
In this context, we introduce \methodName, a new embedded software framework combining novel algorithms to drastically reduce the \ac{RAM} footprint of neural network inference on streams of sensor data time series. More precisely, our contributions are as follows.
%\subsection{Contributions}

\begin{itemize}
    \item We design \methodName, using a \ac{SSM}-based approach, which eliminates redundant computations for sensor data time-series; we also publish its implementation open-source \footnote{see \url{https://github.com/future-proof-iot/tiny-dejavu}}.
    \item We propose a new framework to analyze the causality of compute graph and convert the temporal operator into \ac{SSM} for ultra-low \ac{RAM} streaming process.
    \item We propose a new algorithm to facilitate the overlapping sliding window to further accelerate the inference.
    \item We improve the implementation of global pooling to further decrease \ac{RAM} usage.
    \item We perform reproducible benchmarks with \methodName for various neural network models on common 32-bit microcontroller hardware (Arm Cortex-M). %released preliminary results of deep learning backbones with various structures on \acp{MCU}.
    \item We demonstrate that, compared to prior work (Wavenet, StreamiNNC), we both improve inference performance on overlapping sliding windows by eliminating up to 90\% of the required \ac{RAM} footprint, and extend operator support from pure convolution to the general case. 
\end{itemize}

\added{Note that while this work focuses on computational graph analysis and operator rewriting, \methodName is orthogonal to other resource optimization techniques. In practice, additional optimizations (e.g., quantization, pruning, layer fusion, etc.) can be applied jointly with \methodName.}
\section{Background}

\subsection{Deep Temporal Learning and Sliding Window}

Deep learning approaches for time series modeling have grown rapidly, encompassing a range of architectures. A fundamental step in using such models is framing the learning problem via \textit{sliding windows}: the continuous series is partitioned into overlapping fixed-length segments, each forming an $(input, output)$ training pair with the labeled ground-truth \cite{ndungi2025improving}. This segmentation enables data-driven sequence learning and provides a window-size trade-off: selecting an appropriate size is crucial to capture sufficient context without harming global dynamics. Moreover, the sliding windows using in segmentation are substantially overlapped, to avoid cutting key feature at the edge and to preserve the local, nonstationary behaviors. 

These sliding window inputs are widely used in learning temporal features from data streams, effectively enabling deep models (\acp{RNN}, \acp{CNN}, etc.) applied in the time-series domain. However, to align with the model training phase, the same settings of sliding windows are also put onto the input stream during \textit{model inference}. This leaves a heavy burden for real-time applications and introduces recomputation on the common model architectures, especially on \ac{MCU}-based platforms with long sequence inputs. To alleviate this, this work facilitates the principle of \acp{SSM} to reduce the \ac{RAM} usage of common temporal operators and eliminate the redundant computation of overlapping sliding windows. 
  
\subsection{Temporal Operators as \acp{SSM}}
%are \acp{SSM}}

% TODO: explain more why SSMs, `we found SSMs can well capture the behaviours of streaming process of NN. And from that basis we found optimization points…`

\acp{SSM} are broadly used for modeling system dynamics not only in control engineering, but also in \ac{ML} like hidden markov model and deep sequence \ac{ML} \cite{dao_transformers_2024}. 
Inspired by them, this work identifies the potential of \ac{SSM} in representing common \ac{ML} operators while dealing with time series streaming. Here we briefly give a general discrete form of \acp{SSM} in \cref{eq:ssm}, which maps a temporal input sequence $x_t$ into the hidden states $h_t$ and projects them to an output sequence $y_t$.

\begin{align}
\begin{split}
\label{eq:ssm}
& \mathbf{h_t} = \mathbf{A h_{t-1}} + \mathbf{B} x_t \\ 
& y_t = g(\mathbf{h_t}) + \mathbf{D} x_t
\end{split}
\end{align}
where matrix $\mathbf{A}$ and $\mathbf{B}$ control the internal and external temporal dynamics of the \ac{SSM}, respectively, and the mapping function $g(x)$ determines which operation will be applied on the hidden states.  The term $D x_t$ represents the skip, residual connection between the input and output sequence. We will omit this term in the following discussion because it is trivial to compute.

In previous studies, although not explicitly and formally, the philosophy of \ac{SSM} has in fact guided the design of efficient \ac{ML} operators. One example is Fast WaveNet \cite{paine_fast_2016} (and a recent variant StreamiNNC~\cite{kechris2025don}), which proposed an efficient implementation of the original WaveNet. 
% \added{Another more recent representative, StreamiNNC~\cite{kechris2025don}, shares the same implementation and further investigated the effect of zero-padding.} 
Their implementation turned the stacked 1-D dilated convolution layers into a combination of convolution queues to cache previous computations, which drastically reduced the compute complexity for generating a new single output element over overlap input sequences, from $O(2^L)$ to $O(L)$. This approach is in reality equivalent to cascade \acp{SSM} with $ \mathbf{A}=[0 \ \mathbf{I}; \mathbf{0} \ 0]$, $\mathbf{B} = [0 \ 0 \ \cdots \ 1]^\mathsf{T}$ and $g(x)$ being the convolution operator with only one output element. 

We elaborate this concept with an example. \added{Let two successive convolutional kernel $w_1$ and $w_2$ operate on an input series $x_t$, so the intermediate output $y_{1,t}$ and final output $y_t$ can be described as:}

\begin{align}
\label{eq:conv}
\begin{split}
   y_{1,t} & = \sum_{k=0}^{K_1 - 1} w_1(k) x_{t+k-K_1+1},  \\
   y_t & = \sum_{k=0}^{K_2 - 1} w_2(k) y_{1,t+k-K_2+1}, 
\end{split}
\end{align}

\added{where $K_1, K_2$ are the kernel sizes. We can then rewrite them into vector form:}

\begin{align}
    y_{1,t} & = \mathbf{w_1}^\mathsf{T} \cdot \mathbf{h_{1,t}} \\
   y_t & = \mathbf{w_2}^\mathsf{T} \cdot \mathbf{h_{2,t}}
\end{align}

\added{where $\mathbf{h_{1,t}} = \left[x_{t - K_1+1} \ \dots \ x_{t-1} \ x_t\right]^\mathsf{T}$, $\mathbf{h_{2,t}} = \left[ y_{1, t - K_2+1} \ \dots \ y_{1, t-1} \ y_{1,t} \right]^\mathsf{T}$. We observed that}

\begin{align}
\label{eq:h_conversion}
    \mathbf{h_{1,t}} &= \left(\begin{array}{c}
        x_{t - K_1+1}  \\
        \cdots \\
       x_{t-1} \\
       0
    \end{array}\right) + 
    \left(\begin{array}{c}
        \mathbf{0}  \\
       x_t
    \end{array}\right) \\
    &= \left(\begin{array}{cc}
        0 & \mathbf{I}  \\
        \mathbf{0} & 0
    \end{array}\right) \mathbf{h_{1,t-1}} + 
    \left(\begin{array}{c}
        \mathbf{0}  \\
       1
    \end{array}\right)x_t  \\
    &= \mathbf{A_1} \mathbf{h_{1, t-1}} + \mathbf{B_1} x_t
\end{align}.

\added{Similarly, $\mathbf{h_{2,t}}$ can be rewritten as $\mathbf{h_{2,t}} = \mathbf{A_2} \mathbf{h_{2,t-1}} + \mathbf{B_2} y_{1,t}$, where $\mathbf{A_2}$, $ \mathbf{B_2}$ share the same structure as $\mathbf{A_1}$, $\mathbf{B_1}$, respectively. Now we can transform the convolutional layers (\Cref{eq:conv}) into \acp{SSM}-equivalents:}

\begin{align}
&\left\{
\begin{aligned}
\mathbf{h_{1,t}} &= \mathbf{A_1} \mathbf{h_{1,t-1}} + \mathbf{B_1} x_t\\
y_{1,t} &= \mathbf{w_1}^\mathsf{T} \cdot \mathbf{h_{1,t}}
\end{aligned}
\right.
\\
&\left\{
\begin{aligned}
\mathbf{h_{2,t}} &= \mathbf{A_2} \mathbf{h_{2,t-1}} + \mathbf{B_2} y_{1,t}\\
y_t &= \mathbf{w_2}^\mathsf{T} \cdot \mathbf{h_{2,t}}
\end{aligned}
\right.
.
\end{align} 

% By observing the mathematic form of \ac{SSM}, 
Given the findings above, \acp{SSM} can effectively capture the streaming behavior of neural networks, revealing key opportunities for optimization. From the mathematic form of \ac{SSM}, for each element in output sequence $y_t$, it shows

\begin{itemize}
    \item Constant space complexity, for storing a few hidden states $h_t$;
    \item Constant compute complexity, which requires only the hidden states $h_t$ to compute.
\end{itemize}

These two characteristics yield an interesting feature: The calculation of the entire output sequence $y_t$ requires a constant, small \ac{RAM} usage and the overall computation will scale up linearly alongside the output size.

\section{High-Level Idea \& Formalization}
\label{sec:formalization}

The realization that operators can be seen as SSMs raises two interesting questions: \textbf{(1)} Given an operator on time streaming input, can we find an efficient equivalent \ac{SSM}? \textbf{(2)} Given a series of overlapped input sequences (sliding windows), can we eliminate redundant computation in model inference?

\begin{figure}
    \centering
    \includegraphics[width=0.8\linewidth]{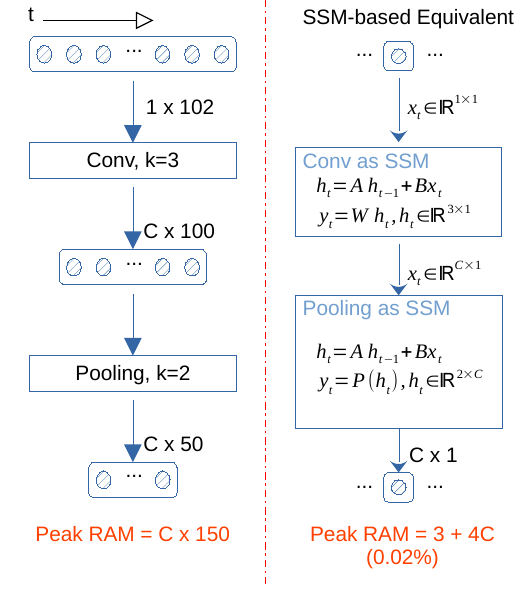}
    \caption{Temporal operators can be expressed as \acp{SSM}, which can drastically reduce peak \ac{RAM} usage. In this example, $k$ and $C$ represent the kernel size and the number of channels, respectively. $P(x)$ denotes the pooling function; $W$ is the kernel weight; $ \mathbf{A}=[0 \ \mathbf{I}; \mathbf{0} \ 0]$; $\mathbf{B} = [0 \ 0 \ \cdots \ 1]^\mathsf{T}$.}
    \label{fig:ops_as_ssm}
\end{figure}

The work we present in this paper seeks to answer these questions by formalizing the equivalence between temporal operators and \acp{SSM}, and proposes a framework to tune the compute graph into \ac{SSM}-based scheme. \Cref{fig:ops_as_ssm} leverages a toy example to show that convolution and pooling operations on time-series can be replaced by \acp{SSM}, which requires only 0.02 \% of the original peak \ac{RAM} usage. We will next further generalize this concept to support various model architectures, whereby greatly reducing \ac{RAM} usage and computation latency on time-series tasks, especially those requiring sliding window inputs. For this, we use the formalization described below.

% are well-suited for modeling the operations of \ac{ML} models on time series, while significantly reducing \ac{RAM} usage during inference.

% - starting point: 
% - Conv/pool/deconv etc. on time series can be seen as SSM
% - Inspired by Fast WaveNet \cite{paine_fast_2016}, we model operators that keeps input causality

\subsection{Sliding Windows on Time Series}

Let \( \mathcal{X} = \{x(t)\}_{t \in \mathbb{N}} \), with \( x(t) \in \mathbb{R}^d \), denote an discrete input data stream. Define the \emph{sliding window operator} \( \mathcal{W}_{l,s} \) as a mapping:

\[
\mathcal{W}_{l,s}(\mathcal{X}) = \{ W_k \}_{k \in \mathbb{N}},
\]

where each window \( W_k \in \mathbb{R}^{l \times d} \) is given by:

\[
W_k = (x_{(k-1)s+1}, x_{(k-1)s+2}, \dots, x_{(k-1)s+l}).
\]

where \( l \in \mathbb{N} \) is the window size, and \( s \in \mathbb{N},  1 \le s \le l \) is the stride. We here define the \textbf{Overlap Rate} of sliding windows as:

\begin{equation}
    r_{overlap} = 1 - \frac{s}{l}
\end{equation}
which refers to how many adjacent windows share common elements. It quantifies redundancy between neighboring windows during operations like feature extraction and segmentation in time series.

\subsection{\ac{SSM}-Equivalent of Temporal Operator}
Considering a input sequence $\mathcal{X} = \{x_t | 1 \le t \le N, t \in \mathbb{N}\}$, we here define a \textit{temporal operator} $\mathcal{T}$ with receptive field size $\tau$ as a function mapping $\mathcal{X}$ to output sequence $\mathcal{Y} = \{y_t\}$ such that:

\begin{equation}
    y_t = \mathcal{T} (\{ x_k | t - \tau + 1\le k \le t \}).
\end{equation}
Considering \emph{stride} $s$ it can be further rewritten as:
\begin{equation}
        y_t = \mathcal{T} (\{ x_k | t \cdot s  - \tau + 1 \le k \le t \cdot s\}).
\end{equation}

Mathematically, this mapping process is \textit{equals to a \ac{SSM} with $\tau$ hidden states and $g(x) = \mathcal{T}(x)$}. Let $h_t = \left[x_{t - \tau +1} \ \dots \ x_{t-1} \ x_t\right]^\mathsf{T}$. We can apply same conversion as in \cref{eq:h_conversion} to get the \ac{SSM}-equivalent:

\begin{align}
&\left\{
\begin{aligned}
\mathbf{h_{t}} &= \mathbf{A} \mathbf{h_{t-1}} + \mathbf{B} x_t\\
y_{t} &= \mathcal{T} (\mathbf{h_{t}})
\end{aligned}
\right.
\end{align}
For stride $s$, this can be handled straightforwardly by defining $y_t=\mathcal{T} (\mathbf{h_{st}})$, where in practice the computation of $\mathcal{T}(\cdot)$ will be skipped for $s-1$ steps.
So, if an operator inside the neural network is considered a temporal operator, we can transform it seamlessly into \ac{SSM}.

Specifically, we call a temporal operator \textit{\ac{GTA}} when $\tau = N$, since it holds a global receptive field on the entire input sequence.

% TODO: add theoretical bound on speed-up / RAM reduction

% \subsection{\ac{GTA}}

\section{Methodology}
\label{sec:methodolgy}

\subsection{Temporal Analysis of Compute Graph}

% Given the definitions of temporal operators, \acp{SSM}, and the \ac{GTA} in \cref{sec:formalization}, 

At first, \methodName examines the temporal characteristics within a typical neural network designed for time-series modeling. These networks comprise layers with varying receptive fields, some operating locally (small $\tau$) and others globally ($\tau = N$).

We identify the \ac{GTA} as a natural boundary that partitions the network into two functional regions:
\begin{itemize}
\item \textbf{\ac{SSM}-subgraph}: Contains only local or causal temporal operators, which are amenable to efficient streaming execution via their \ac{SSM} equivalents.
\item \textbf{\ac{GTA}-subgraph}: Begins at the \ac{GTA} and includes all subsequent acausal or global layers.
\end{itemize}

\begin{figure}[ht]
    \centering
    \includegraphics[width=0.8\linewidth]{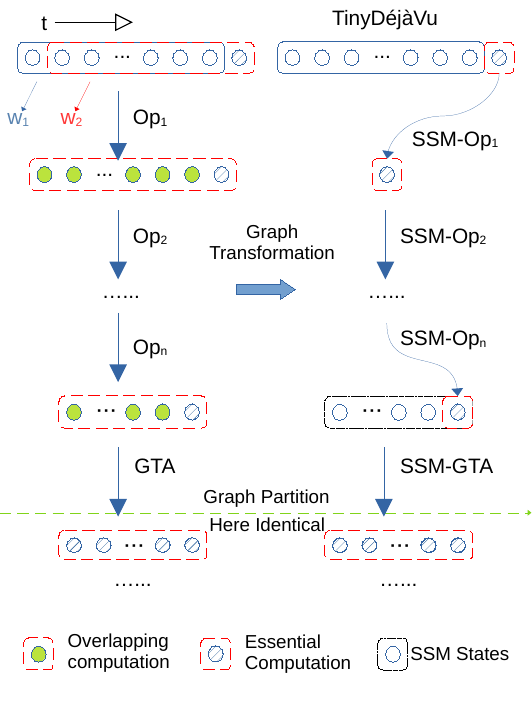}
    \caption{Graph transformation of \methodName. $w_1$ and $w_2$ denote two consecutive, overlapping sliding windows. GTA: Global Temporal Aggregator, Op: Operator.}
    \label{fig:graph_as_cascade_ssm}
\end{figure}

As shown in \cref{fig:graph_as_cascade_ssm}, the neural network is partitioned into two subgraphs, with the \ac{GTA} operator serving as the boundary. Each new data point introduced by sliding windows has a global effect on all layers within the \ac{GTA}-subgraph, whereas in the \ac{SSM}-subgraph, only a small subset of features needs to be computed and stored. This indicates potential savings in both \ac{RAM} and computational resources. As the overlap rate increases, a larger portion of the computation becomes redundant due to overlapping inputs in temporal operators. Transforming these components into \acp{SSM} helps relieve this recomputation issue, as \acp{SSM} process only newly arrived data from preceding layers.

We here give the temporal parameters of common operators in \cref{tab:temp_params}. As discussed in \cref{sec:formalization}, these operators can be transformed into \acp{SSM} with $\tau$ hidden states and $ \mathbf{A}=[0 \ \mathbf{I}; \mathbf{0} \ 0]$, $\mathbf{B} = [0 \ 0 \ \cdots \ 1]^\mathsf{T}$. This transformation reduces \ac{RAM} usage to $\tau / N$ of that required by the original operators.

\begin{table}[ht]
    \caption{Temporal parameters of common operators. $k$: kernel size, $d$: dilation, $N$: length of input, $W$: weights, $b$: bias.}
    \label{tab:temp_params}
    \centering
    \begin{tabular}{llll}
    \toprule
      Operator   & $\tau$ & $\mathcal{T}(x)$ & \ac{RAM} $\downarrow$ \\\midrule
    Conv & $(k-1)d + 1$ & $W * x + b$ & \multirow{2}*{$\tau/N$}\\ 
    Pooling & $k$ & $\text{Max}(x)$ or $\text{Avg}(x)$ &\\\midrule
    Dense$^*$ & N & $Wx + b$ & - \\
    Attention$^*$ & N &  $\text{Atten}(Q(x), K(x), V(x))$ & - \\
    \bottomrule
    \end{tabular}
    \\
    \footnotesize{$^*$Global Temporal Aggregator.}
\end{table}

Specifically, A dense layer is treated as a global temporal aggregator only when it mixes values across the temporal dimension, for example after flattening a sequence representation into a single vector. In contrast, a dense layer applied independently to each time step performs channel mixing only and is treated as a temporal operator with $\tau=1$.

It is noted that \ac{RNN} and its variants such as \ac{LSTM} and \ac{GRU} are \acp{SSM} with $\tau=1$ by design, thus \methodName will automatically keep them untouched during graph transformation. For \textit{strides} behavior of operators, we explain the implementation details in \cref{sec:implementation}.

% This partitioning allows us to streamline the compute graph, reducing memory consumption during inference while maintaining temporal fidelity.

\subsection{Deep Sliding Window on \ac{GTA}-Subgraph}

As discussed in \cref{sec:formalization}, streaming applications commonly adopt overlapping sliding windows to improve temporal resolution and output stability. As shown in \cref{fig:graph_as_cascade_ssm}, the neural network takes two sliding windows, $w_1$ and $w_2$, as consecutive inputs from the data stream.
% Let each window have length $L$ and stride $s$; the overlap rate $\mathcal{O}$ has been defined in \cref{sec:formalization}.

Standard approaches recompute the full model for each window, leading to high redundant computation. Instead, \methodName eliminates all redundant computation of the temporal operators by replacing them by the equivalent \acp{SSM}. The \ac{GTA} is also transformed into \ac{SSM} form, which responsible for:
\begin{enumerate}
\item Cache historical intermediate outputs from the SSM-subgraph as hidden states.
\item Determine which hidden states should be replaced / updated, according to new inputs introduced by the current sliding window.
\item Feed the resulting updated hidden states as input feature maps into the GTA-subgraph.
\end{enumerate}

To further reduce compute latency, \methodName performs temporal analysis to determine how many hidden states must be updated in response to new data introduced by consecutive sliding windows. This value is then used to set the stride of the \ac{GTA}, ensuring that the \ac{GTA} initiates computation only when all relevant new input features from the \ac{SSM}-subgraph have been received.

% Because the SSM-subgraph is both causal and compossible, this design supports efficient reuse across overlapping windows without recomputation. The overall  compute cost becomes proportional to the overlap rate of sliding window, as define in \cref{sec:formalization}.

\subsection{Transform Neural Network as Cascade \acp{SSM}}

After extracting operator parameters during temporal analysis, all temporal operators in the \ac{SSM}-subgraph are transformed into the corresponding \acp{SSM} to support efficient sequential inference. By expressing each operator as a time-recursive update of internal states, we avoid buffering large input windows and instead maintain only the current hidden state.

This transformation enables the network to process input tokens incrementally, making it highly suitable for resource-constrained or low-latency scenarios such as online streaming. The result is a cascade of compact, recurrent \ac{SSM} layers that collectively mimic the behavior of the original model with significantly reduced memory requirements.

% TODO: add more explaination on what is preheat, what is Streaming phases

During model inference, the computation of \methodName is split into two stages:

\begin{itemize}
    \item \textbf{Preheat}: Full computation of the first input window, where all initial hidden states are calculated and cached;
    \item \textbf{Streaming}: After \textit{preheat} stage, the model starts receiving new data from the consecutive sliding windows, where only the new data, instead of whole window,  will be processed.
\end{itemize}

It is noted that the preheat stage corresponds to the vanilla non-\ac{SSM} setup, where all consecutive overlapping sliding windows are recomputed from the beginning.

\section{\methodName Implementation Details}
\label{sec:implementation}

We have implemented the \methodName mechanism on top of Pytorch v2.3.0 \cite{paszke2019pytorch} and microTVM v0.16.0 \cite{chen2018tvm}. We used the Pytorch frontend to conduct temporal analysis on original models. Then we conveyed the analysis results into TVM frontend to rewrite the compute graph and generate low-level routines of \acp{SSM} to fit the settings. We leveraged RIOT-ML \cite{huang2024riot-ml} to benchmark the models (transform into C code by microTVM).

\subsection{\acp{SSM} Implementation}
% {\noindent\bf\acp{SSM} Implementation --}
Provided all equivalent \acp{SSM} except global pooling sharing the same  $ \mathbf{A}=[0 \ \mathbf{I}; \mathbf{0} \ 0]$, $\mathbf{B} = [0 \ 0 \ \cdots \ 1]^\mathsf{T}$ structure, we implement them efficiently using a circular buffer to eliminate unnecessary computation. In \cref{alg:ssm}, we provide implementation details with consideration of the original operators' \textit{stride} settings. The circular buffer has a fixed length of \(\tau\), such that outdated hidden states are replaced by the current input data point \(x_t\). The function \textit{WaitForNextInput()} pauses the execution of the compute graph when the stride condition is not satisfied.

\begin{algorithm}[tb]
\caption{Implementation of \acp{SSM}}
\label{alg:ssm}
\begin{algorithmic}[1] %[1] enables line numbers
\item[\textbf{Input}:] Data point $x_t$, time index $t$

\item[\textbf{Property}:] number of hidden states $\tau$, circular buffer $B$ with length $\tau$, temporal operator $\mathcal{T}(x)$, \ac{SSM} stride $s$ 

\item[\textbf{Output}:] $y_t$

\STATE $B$.insert($x_t$)

\IF {$t \% s == 0 $}
\STATE Let $h_t$ = B.flatten()
\STATE $y_t = \mathcal{T}(h_t)$
\STATE \textbf{return} $y_t$
\ELSE
\STATE WaitForNextInput()
\ENDIF

\end{algorithmic}
\end{algorithm}

% TODO: add clarification that all graph analysis / transform / model compilation happens on PC (and very fast), not thing needs to be done on end-devices.

\subsection{Global Pooling Optimization}
\begin{figure}[ht]
    \centering
    \includegraphics[width=0.8\linewidth]{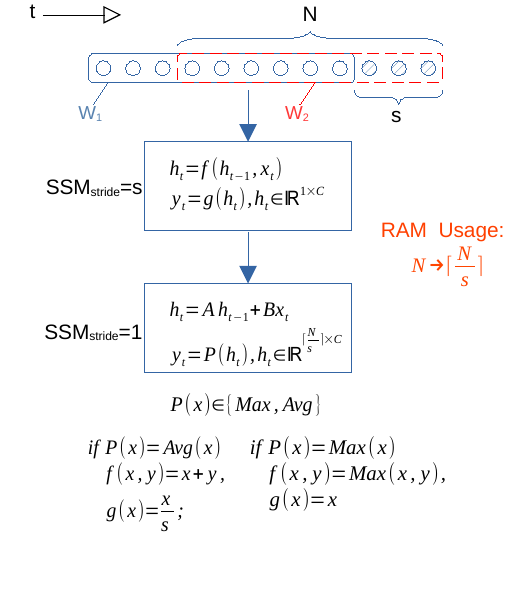}
    \caption{Equivalent \ac{SSM} of Global Pooling. $N$: window size, $s$: window stride.}
    \label{fig:global_pooling}
\end{figure}

Here, we discuss a special case of the \ac{GTA} operator: global pooling. When applied to sliding windows, global pooling can be replaced by a cascade of smaller \acp{SSM}, enabling significant reductions in \ac{RAM} usage. Inspired by msf-CNN~\cite{huang2025msf}, which demonstrated that pooling outputs can be computed iteratively, this work derives an equivalent \ac{SSM} representation, illustrated in \cref{fig:global_pooling}. Rather than maintaining a large buffer of hidden states to span the full global receptive field \(\tau = N\), we employ a two-stage approach: a first \ac{SSM} with a single hidden state performs partial aggregation over input chunks of size \(s\), followed by a second \ac{SSM} acting as a global aggregator with \(\left \lceil{N/s}\right \rceil \) hidden states. 
If $N$ is not divisible by $s$, a counter is used to track whether $N$ elements have already passed through the first \ac{SSM} in the current iteration. Once this condition is met, the second \ac{SSM} is triggered to compute the final result. The denominator of $g(x)$ is adjusted accordingly to match the actual pooling behavior.
This strategy reduces the overall \ac{RAM} complexity from \(\mathcal{O}(N)\) to \(\mathcal{O}(\left \lceil{N/s}\right \rceil)\), offering a more memory-efficient solution for streaming inference.

% {\noindent\bf Global Pooling Optimization --}
% Here we discuss a special \ac{GTA}, global pooling, which can be replaced by cascade, smaller \acp{SSM} when confronting sliding windows, to further reduce the \ac{RAM} usage. Inspired by msf-CNN \cite{huang2025msf}, which demonstrated that the outputs of pooling layer can be computed iteratively, this work gives its equivalent \ac{SSM} form, as shown in \cref{fig:global_pooling}. Instead of using a huge buffer storing hidden states to match the global receptive field $\tau=N$, we here use a \ac{SSM} with one hidden state to partially pre-aggregate the input data with size $s$, followed by another \ac{SSM} as global aggregator with $l/s$ hidden states. So we further squeeze the \ac{RAM} usage from $\mathcal{O}(l)$ to $\mathcal{O}(l/s)$. 

\subsection{\textit{Optional:} BF16 Optimization}
% {\noindent\bf BF16 Optimization --}

% TODO: could be confused, clarify that we provide bf16 as option, not hard-coded inside our implementation.

To further reduce \ac{RAM}  usage, we adopt \textit{BF16 (brain floating point with 16 bits)} \cite{kalamkar2019study} precision as an option to store hidden states throughout the \ac{SSM}-subgraph. BF16 provides a favorable trade-off between dynamic range and bit width, enabling a 2$\times$ reduction in \ac{RAM} usage of hidden states compared to FP32. User can choose either full precision (FP32) or BF16 to have a better trade-off between \ac{RAM} usage and model accuracy.

In our implementation, all state storage is kept in BF16 during inference. Casting to FP32 occurs only for precision-critical computations, e.g. the computation of \ac{SSM} output $y_t$, leveraging mixed-precision hardware support. This yields minimal conversion overhead while preserving model accuracy.
\section{Correctness and Task-level Evaluation}
To validate the correctness of the graph transformation and operator rewriting, we computed both the relative and absolute deviations between the original (Vanilla) model and the \methodName-transformed model using randomly generated inputs and weights. This evaluation was performed at both the individual operator level and the end-to-end model level. The results show that all deviations are below $10^{-8}$, thereby confirming numerical correctness.

Moreover, we provide task-level evaluation to further confirm the transformation correctness. \Cref{tbl:task-perf} shows that no discrepancy was found between Vanilla and \methodName. \Cref{tbl:task-perf}  also includes results obtained under BF16 optimization. In this setting, a slight performance drop can be observed, indicating that retraining may be necessary for applications requiring high predictive accuracy.
% \added{in Appendix \ref{app:task_eval}, showing that no task performance drop was found between Vanilla and \methodName. }

\begin{table*}[]
\caption{Transformation correctness validated by task performance.}
\label{tbl:task-perf}
\centering
\begin{tabular}{lllllll}
\toprule
Methods         & CET-S* & ResTCN** & TEMPONet* & TC-CNN* & TC-TFM* & WaveNet** \\\midrule
Vanilla         & 69.8   & 2.97     &  60.9        & 94.3    & 93.7    &   12.6      \\\midrule
\methodName \\
 FP32 & 69.8   & 2.97    &  60.9        & 94.3    & 93.7    &    12.6     \\
 BF16 & 65.3   & 3.40     &  55.4        & 91.2    & 90.5    &   16.5      \\
\bottomrule
\end{tabular}
\\ $*$ Classification tasks performance was measured in terms of accuracy in percentage (higher is better).
\\ $**$ Generative tasks performances was measured in terms of negative log-likelihood loss (lower is better).
\end{table*}

{\noindent\bf Impact on Model Accuracy of BF16 --}
To evaluate the impact on accuracy of using BF16 as hidden states storage, we calculated the relative root-mean-square error (RMSE) of outputs between FP32 and BF16 variants. Though the RMSE ranges from 1\% to 3\% for all models we used in this work, the task-level accuracy drop is considered notable. On the other hand, this work focuses on reduction of \ac{RAM} usage and optimization of compute latency of streaming process. To ensure a negligible accuracy drop, We recommend re-trained the model under BF16 scheme for better performance before enabling BF16 support.

\section{Experiments and Discussion}

In this section, we present results on experiments running \methodName on the host PC and \ac{MCU}, aiming to validate both the correctness of our optimization strategies and their versatility when applied on diverse model architectures.

More concretely, we measured peak \ac{RAM} usage and compute latency on sliding window inputs with various overlap rates based on the optimization technique described in \cref{sec:methodolgy} and \cref{sec:implementation}, as reported in the following. We define peak \ac{RAM} as usual in the literature, i.e., the highest total \ac{RAM} allocated currently, at any single point in time, during the execution of inference. 

\textit{ML Model —} To explore the generality and capability of \methodName, here we choose models among a zoo of diverse hybrid architectures and tasks. TC-CNN \cite{huang2024tinychirp}, TEMPONet \cite{zanghieri_robust_2020} and ResTCN \cite{bai_empirical_2018} use convolutions and pooling as feature extractors, followed by dense blocks for classification or generation. In contrast, CET-S \cite{rohr_transformer_2023} and TC-TFM \cite{huang2024tinychirp} used Transformer blocks as the final processing stage. Input window sizes were kept consistent with those used in the original studies. Additionally, we analyze the impact of varying overlap rates on computation latency. \emph{Vanilla} refers to the untransformed models, which were compiled and optimized by microTVM with the best memory schedule. All models, unless explicitly marked, were benchmarked under FP32 precision.

In addition, we examined the WaveNet \cite{oord_wavenet_2016}, a special generation model purely composed with temporal, dilated 1-D convolutions.
% , to quickly validate the \methodName's correctness on transforming operators into cascade \acp{SSM}. 
To make it feasible to benchmark on the target \ac{MCU}, we tailored the model to meet the memory constraint, cutting down 75\% of the layers and 90\% of the residual channels, named as \textit{TinyWaveNet} in this paper.
% We let the model to generate 10000 data points and averaged the compute latency of each generated sample. We also measured the compute overhead of using BF16 format for hidden states storage.

\textit{Reproducibility —} Though we carried out experiments on a specific \ac{MCU}, \methodName itself generates platform-agnostic C-code, thus can actually run on any CPUs (such as x86 and Cortex-A) or on other \acp{MCU} (ESP32, RISC-V and other Cortex-M). To support reproducibility, we have open-sourced the implementation of \methodName along with all the models used in our experiments. Random initialization settings are irrelevant, as all experiments are deterministic.

It is noted that all graph transformations and code generation are performed on the host PC and do not consume any resources on the \ac{MCU}. The analysis of each operator and its transformation to \ac{SSM} are with complexity of $\mathcal{O} (1)$. Thus, the computational overhead on the host PC is negligible, particularly for models with fewer than 1000 layers, which are suitable for deployment on \ac{MCU}-based devices.

\subsection{Impact on \ac{RAM} footprint}
\begin{figure}[ht]
    \centering
    \includegraphics[width=0.8\linewidth]{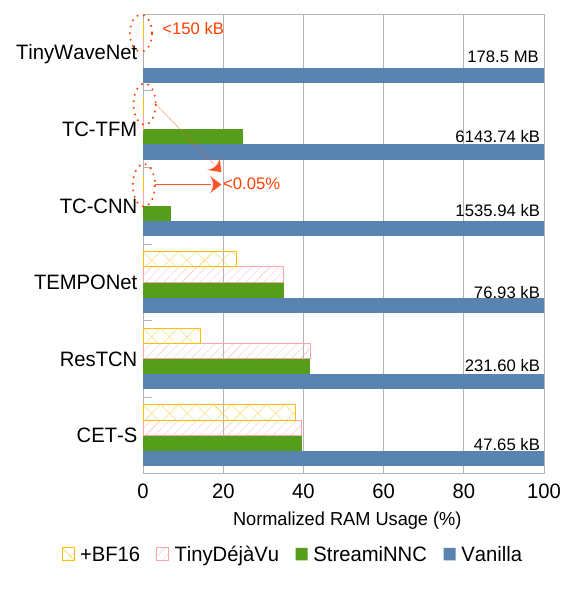}
    \caption{RAM Usage: Vanilla vs. \methodName. All results are normalized under Vanilla models. The numbers on the right side denote the peak RAM usage of vanilla models.}
    \label{fig:ram-usage}
\end{figure}

\ac{RAM} footprints are obtained from microTVM’s static memory planner, and thus remain identical regardless of the target deploy.
\Cref{fig:ram-usage} presents a great potential of \methodName in saving \ac{RAM} usage, achieved at least 60\% \ac{RAM} usage reduction compared to the vanilla. 
% In particular when facing a long input sequence, it took only 0.08\% \ac{RAM} usage of the vanilla TinyWaveNet. 
% As depicted in \cref{fig:ram-usage}, \methodName achieved at least 60\% \ac{RAM} usage reduction compared to the vanilla.
The most significant savings were observed with TinyWaveNet, TC-TFM and TC-CNN, which exhibited exceptional efficiency gains compared to the state-of-the-art (StreamiNNC). These models feature relatively small receptive fields compared to their large input window sizes (e.g., 48K vs. 3 on TC-TCM/CNN, 10K vs. 512 on TinyWaveNet). When facing a long input sequence, the vanilla model produces long-sequence intermediate activations as well, which occupies a large amount of \ac{RAM}; On the contrary, \methodName decouples \ac{RAM} usage with the input length by employing \acp{SSM} with fixed receptive fields, which are vastly smaller than input length and thus consume smaller buffer spaces. This offers substantial optimization opportunities for \methodName through the conversion of temporal operators into \acp{SSM}. 

Additionally, enabling BF16 further contributes to RAM savings. An exception was observed with CET-S: upon deeper investigation, we found that its Transformer block contains an attention layer modeled as a \ac{GTA}, which dominates overall memory consumption. As a result, BF16 had limited impact on \ac{RAM} reduction for this model compared to the others.

% \subsection{Pilot Study: \methodName on WaveNet}
\subsection{Pilot Study: Inference Latency on Host PC}
Since certain vanilla, untransformed models would cause out-of-memory (OOM) on microcontrollers, we first performed benchmarking on a host PC instead. The PC is equipped with an Intel Core i7-1165G7 CPU at 2.80 GHz, running Ubuntu 24.04.2 LTS on WSL2.

As shown in \cref{tbl:latency_on_host}, \methodName delivers significant acceleration during the streaming stage compared to untransformed vanilla models. With an overlap rate of $0.5$, it achieves at least a $2 \times$ reduction in latency, and up to $6 \times$ in the best cases, which is particularly important for time-sensitive systems. We observe a slight increase in latency during the preheat stage relative to vanilla models, even though both are mathematically equivalent and share the same arithmetic routines. However, because the preheat stage is executed only once to initialize the \acp{SSM}, this overhead is negligible in the context of streaming inference.

\begin{table*}[h]
\caption{Compute latency in ms measured on the host PC. }
\label{tbl:latency_on_host}
\centering
\begin{tabular}{lllllll}
\toprule
Methods                 & CET-S & ResTCN & TEMPONet & TC-CNN & TC-TFM & TinyWaveNet \\\midrule
Vanilla  & $1.3 \pm 0.02$   & $53.7 \pm 1.57$   & $15.3 \pm 0.59$     & $7.2 \pm 0.34$    & $6.1 \pm 0.46$    & $61.5 \pm 1.86$    \\\midrule
\methodName \\
Preheat                & $1.5 \pm 0.09$   & $55.3 \pm 1.15$   & $15.8 \pm 0.43$     & $8.2 \pm 0.37$    & $9.3 \pm 0.51$    & $70.6 \pm 1.43$    \\
Streaming, r=0.5       & $0.8 \pm 0.01$   & $8.7 \pm 0.18$    & $4.0 \pm 0.10$      & $4.1 \pm 0.14$   & $4.9 \pm 0.19$    & $34.5 \pm 0.65$    \\\bottomrule
\end{tabular}
\\
$r$: overlap rate of sliding window. All reported values are presented as mean $\pm$ standard deviation.
\end{table*}

\subsection{Compact Study: Inference Latency on STM32F7 \ac{MCU}}

We further conducted our experiments on the STM32F746NG \ac{MCU} with 340 kB \ac{RAM}, 1 MB Flash and an Arm Cortex-M7 CPU running on 216 MHz core frequency with instruction and data cache enabled. Since certain vanilla models are infeasible to deploy on the target \ac{MCU} due to their excessive memory requirements (up to 180 MB of \ac{RAM}), we benchmark \methodName here without comparison to the vanilla models. We measure streaming inference latency across overlap rates ranging from 0\% to 90\%, with a step size of 10\%.

\begin{figure}[ht]
    \centering
    \includegraphics[width=0.8\linewidth]{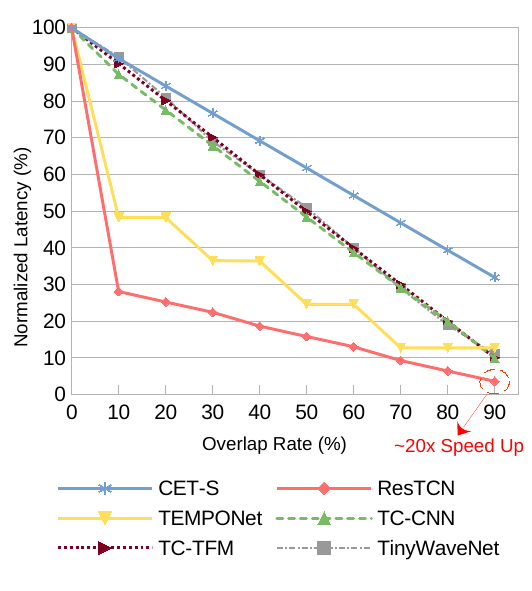}
    \caption{Compute latency during Streaming stage measured on stm32f746g-disco board with different overlap rates of sliding windows. All results are normalized under baseline (preheat) latency in \cref{tab:latency-baseline}.}
    \label{fig:latency-vs-overlap-rate}
\end{figure}

{\noindent\bf Impact of Overlap Rate --}
\Cref{fig:latency-vs-overlap-rate} illustrates the impact of overlap rate on latency during the streaming stage. All latency values are normalized with respect to the Preheat stage baselines reported in \cref{tab:latency-baseline}. The results show a clear linear decrease in latency as the overlap rate increases, proving \methodName's effectiveness in eliminating redundant computation across sliding windows.
However, two outliers, ResTCN and TEMPONet, present a different pattern. Their latency curves show a significant drop at the 10\% overlap rate, followed by a more gradual decline. This behavior can be traced back to their deeper network architectures, which allow more historical information to be retained within the \acp{SSM}. Even though lower overlap rates introduce more new data, a substantial portion of the hidden states -- particularly in deeper layers -- remains unchanged. As a result, the computational cost during the streaming stage remains relatively low despite increased input variation.

% \Cref{fig:latency-vs-overlap-rate} reports the influence of overlap rate during streaming stage. The latencies were normalized under those of preheat stage in \cref{tab:latency-baseline}. It clearly shows that the latency is decreased linearly alongside the increment of overlap rate, which again validates the ability of \methodName on eliminating re-computation of sliding windows. However, we observes two outliers, ResTCN and TEMPONet, whose latencies' curves have a decent drop on 10\% overlap rate, then become gradual afterwards. The reason is they have much deeper structure and thus more historical information are preserved inside the \acp{SSM}. Though more new data are brought in due to lower overlap rate, most of the hidden states deepened inside the neural network remains unchanged, leading to low compute latency during streaming stage.

\begin{table*}[ht]
\caption{Compute latency in ms of baseline (\textit{preheat}) measured on stm32f746g-disco board.}
\label{tab:latency-baseline}
    \centering
    
    \begin{tabular}{llllll}
    \toprule
CET-S & ResTCN & TEMPONet & TC-CNN & TC-TFM & TinyWaveNet \\\midrule
$59.2 \pm 0.02$  & $1669.6 \pm 1.88$ & $521.5 \pm 0.30$    & $295.5 \pm 4.63$  & $345.5 \pm 0.56$  & $2776.1 \pm 2.10$\\
\bottomrule
\end{tabular}
\\
All reported values are presented as mean $\pm$ standard deviation.

\end{table*}

In summary, our experiments demonstrate that \methodName effectively optimizes resource usage across diverse temporal models by adapting to user-specified sliding window overlap rates, which control the degree of redundant computation. This allows users to generate optimized model code tailored to specific industrial and real-time streaming applications.

Although direct power measurements are not reported in this work, \methodName reduces both peak \ac{RAM} usage and inference latency. It allows devices to stay longer in low-power states and to reduce memory accesses and data movement, which may further decrease power consumption during inference. 
\section{Use Case: Bird Sound Detection in UK}

Beyond benchmarking on various streaming models, we apply \methodName with several real-world scenarios. One representative use case is bioacoustic monitoring on low-power sensors. These battery-powered devices were deployed in rural areas of the UK for an entire season to monitor common corn bunting songs. Building on TinyChirp \cite{huang2024tinychirp}, we further optimized TC-CNN and TC-TFM using \methodName and deployed them on sensors equipped with an nRF52840 \ac{MCU}. Compared to TinyChirp, \methodName reduces \ac{RAM} usage by $75 \times$ (approximately 1 kB vs. 75 kB) and achieves 60 \% reduction in inference latency, which is a critical improvement for such low-power deployment scenarios. These numbers validate our method in real life, beyond lab benchmarks.

\section{Related Work}

\subsection{Machine Learning Compilers}
Compilers such as \ac{TVM} \cite{chen2018tvm}, IREE \cite{The_IREE_Authors_IREE_2019}, FlexTensor \cite{zheng2020flextensor}, and Buddy \cite{zhang2023compiler} provide automated transpilation and compilation pipelines for models developed in major \ac{ML} frameworks, including TensorFlow and PyTorch. As a lightweight extension of TVM, microTVM offers low-level optimizations and runtime support tailored to a variety of processing units, including many microcontroller architectures. Prior work such as RIOT-ML \cite{huang2024riot-ml} integrates a lightweight general-purpose OS with microTVM to support end-to-end deployment of ML models and operator implementations across diverse \acp{MCU}. Similarly, \methodName builds upon microTVM, using it both as a front-end importer for model files and as a code generator targeting resource-constrained platforms.
% However, none of the above tools provide CNN fusion optimization mechanisms, in contrast to msf-CNN.
\subsection{Deep Learning for Time Series Modeling}
Deep Learning has been widely applied to time series analysis \cite{lim2021time, gamboa2017deep, ismail2019deep, morid2023time}, especially in real-time applications such as voice generation, healthcare, anomaly detection and forecasting.

Recurrent architectures \cite{jai.2024.054314} were the pioneering deep models applied to time series because of their natural fit to sequential data. Vanilla \acp{RNN} are designed for sequences, but suffer from vanishing gradients, so gated variants like \ac{LSTM} and \ac{GRU} became standard to capture longer-range dynamics.

\acp{CNN} have also been widely applied to time series. \ac{TCN} can efficiently capture local temporal patterns by sliding filters over the input windows. A pioneer for voice generation, WaveNet \cite{oord_wavenet_2016}, introduced temporal and dilated convolutions stacked together to model temporal dependencies without recurrence, provided a very large receptive field. TEMPONet \cite{zanghieri_robust_2020}, ResTCN \cite{bai_empirical_2018} and TinyChirp \cite{huang2024tinychirp} present the ability of temporal convolutions in gesture recognition, sequence modeling and sound detection, respectively.
% Building on this, the Temporal Convolutional Network (TCN) uses stacked dilated convolutional layers with increasing dilation factors, achieving a very large receptive field.
\cite{bai_empirical_2018} suggests that deep convolutional designs can capture long-term dependencies at least as well as \acp{RNN}.

More recently, Transformers \cite{vaswani2017attention} have been adapted for time series forecasting, leveraging self-attention to capture long-range interactions. Transformers shows decent potential at capturing global dependencies in sequential data \cite{wen2022transformers}. Many Transformer variants have been proposed for time series: CET \cite{rohr_transformer_2023} combined ResNet and transformer blocks to achieve promising scores for predicting post-cardiopulmonary resuscitation outcome based on \ac{EEG}; TinyChirp \cite{huang2024tinychirp} used a single-head transformer as the classifier processing the temporal features extracted from a long audio sequence with 48k data points. 

In summary, recent works in deep time series modeling span classical recurrent networks (RNN/LSTM/GRU), convolutional architectures (\acp{CNN} and \acp{TCN}), and attention-based transformers, often in hybrid combinations.

\subsection{Optimization for Time Series Inference}
Several studies have leveraged the aforementioned causality to avoid redundant computation and improve the efficiency of model inference on time series.
One well-known example is Fast WaveNet \cite{paine_fast_2016}, which caches intermediate feature maps in voice generation to reduce per-sample complexity from $\mathcal{O}(2^L)$ to $\mathcal{O}(L)$ for WaveNet with $l$ dilated convolutions. 
\cite{burrello_tcn_2021} developed a specialized kernel library for \acp{TCN} on low-power devices, rewriting 1-D convolution kernels to exploit causal structure and significantly reduce latency and energy usage.
\cite{kechris2025don} introduced StreamiNNC, a framework for streaming \ac{CNN} inference that exploits convolutional shift-invariance to skip redundant operations in overlapping windows by caching previous outputs and computing only for newly arrived inputs.
\cite{mudraje20251} proposed a lightweight inference engine for interleaved 1D-\ac{CNN} execution on microcontrollers, which interleaves data acquisition with incremental convolution using ring buffers, reducing both latency and memory footprint compared to standard deployment pipelines.

However, existing approaches either focus exclusively on 1D convolutions or lack support for overlapping sliding windows. None of them above provides an end-to-end code generation pipeline for diverse, hybrid architectures targeting universal \acp{MCU}, in contrast to \methodName. 
% To address these limitations, \methodName formalizes temporal operators as \acp{SSM} and introduces a framework for evaluating the impact of overlapping sliding windows on resource usage. Additionally, it includes a model transpiler that supports deployment across general CPU-based platforms.

% However, all of them either focused purely on 1-D convolutions, or lacked support for overlapping sliding windows. None of them proposed a end-to-end codegen pipeline for diverse, hybrid architectures targeting on \acp{MCU}. To fill in these gaps, \methodName formalized the temporal operators as \acp{SSM}, and proposed a framework to evaluate the impact of overlapping sliding windows on resource usage, as well a model transpiler supporting general CPU-based platform.

% \cite{burrello_tcn_2021} rewrites the kernel of \ac{Conv1D} to accelerate inference, by facilitating the causality.

% Fast WaveNet \cite{paine_fast_2016} optimizing WaveNet caches elements of feature maps of each \ac{Conv1D} layers, to accelerate (one) voice sample generation (TTS). Fast WaveNet only focus on \ac{Conv1D} and generative task, did not consider other operators and tasks. Fast Wavenet not consider (partially overlapped) sliding window inside the neural network, which causes recompute issue.
\section{Conclusion}

Embedded intelligence is deployed on cheap, always-on devices using small artificial neural networks to continuously perform inference on time series of sensor data. These devices use microcontroller-based hardware, whose smaller energy consumption and tinier price tag are determined by peak \ac{RAM} usage. Targeting this segment, in this paper, we designed \methodName, a memory- and compute-efficient framework for time-series inference transforming temporal operators into \acp{SSM}, %isolating the global aggregator, 
and leveraging streaming-aware optimizations such as sliding window reuse and mixed-precision arithmetic.
\methodName can reduce peak \ac{RAM} use by up to 90\% compared to state-of-the-art without affecting inference accuracy and can achieve up to $20 \times$ speed up by eliminating redundant computation on data stream. We published an open source implementation of \methodName which is portable on most commercially available microcontroller boards. This makes \methodName an interesting tool for the design of drastically more energy-efficient long-sequence processing of sensor data time series on extremely resource-constrained devices.

% \fi
% \IEEEpeerreviewmaketitle



\bibliographystyle{IEEEtran}
\bibliography{aaai2026}
\balance
%

% \input{chp/99-appendix}

\end{document}